\documentclass[a4paper]{easychair}
\usepackage{comment}
\usepackage{amsmath}
\usepackage{bbold}
\usepackage{comment}
\usepackage{booktabs}
\usepackage{multirow}
\usepackage{graphicx}
\usepackage{subfigure}
\usepackage[usenames]{xcolor}
\usepackage{colortbl}
\usepackage{array}
\newcolumntype{M}{>{\centering\arraybackslash}m{\dimexpr.25\linewidth-2\tabcolsep}}
\usepackage[inline]{enumitem}

\newcommand{\newcite}{\cite}

\newcommand{\word}[1]{\emph{#1}}
\newcommand{\tech}[1]{\emph{#1}}

\newcommand{\Secref}[1]{Section~\ref{#1}}
\newcommand{\secref}[1]{Section~\ref{#1}}

\newcommand{\Figref}[1]{Figure~\ref{#1}}
\newcommand{\figref}[1]{Figure~\ref{#1}}

\newcommand{\tabref}[1]{Table~\ref{#1}}
\newcommand{\pararef}[1]{\textbf{#1}}

\newcommand{\given}{\mid}

\newcommand{\tuple}[1]{\left<#1\right>}
\newcommand{\cond}[1]{\emph{#1}}

\newcommand{\emptystring}{\emptyset}
\newcommand{\uttattr}[1]{$[$\textit{#1}$]$}
\newcommand{\refattr}[1]{\textsc{#1}}
\newcommand{\featpair}[2]{\refattr{#2} $\wedge$ \uttattr{#1}}
\newcommand{\uttheader}[1]{\rotatebox{90}{#1}}

\newcommand{\refheader}[1]{#1}

\newcommand{\context}{c}
\newcommand{\States}{T}
\newcommand{\state}{t}
\newcommand{\Lex}{\mathcal{L}}
\newcommand{\Messages}{M}
\newcommand{\msg}{m}
\newcommand{\Costs}{C}
\newcommand{\StatePrior}{P}

\newcommand{\speakerZero}{s_{0}}
\newcommand{\speakerOne}{s_{1}}
\newcommand{\listenerOne}{l_{1}}

\newcommand{\expect}{\mathbb{E}}

\newcommand{\best}[1]{\textbf{#1}}
\newcommand{\close}[1]{\textbf{\textit{#1}}}

\newcommand{\graycell}[1]{{\cellcolor[gray]{.8}#1}}

\newcommand{\sigweight}[1]{\textbf{#1}}

\newcommand{\marginnote}[1]{}

\renewcommand{\marginnote}[1]{}

\title{Learning in the Rational Speech Acts Model}
\titlerunning{Learning in the Rational Speech Acts Model}

\author{Will Monroe \and Christopher Potts}

\institute{Stanford University, California, U.S.A. \\
\email{wmonroe4@cs.stanford.edu}, \email{cgpotts@stanford.edu}}

\authorrunning{Monroe and Potts}

\begin{document}

\maketitle

\marginnote{Tweaked the wording on the ``activation function'' sentence.}

\begin{abstract}
  The Rational Speech Acts (RSA) model treats language use as a
  recursive process in which probabilistic speaker and listener agents
  reason about each other's intentions to enrich the literal semantics
  of their language along broadly Gricean lines. RSA has been shown to
  capture many kinds of conversational implicature, 
  but it has been criticized as an unrealistic model of speakers, 
  and it has so far required the manual specification of a semantic lexicon,
  preventing its use in natural language processing applications that
  learn lexical knowledge from data.
  We address these concerns by showing how to define and optimize
  a trained statistical classifier that uses the intermediate agents
  of RSA as hidden layers of representation forming a non-linear  
  activation function. This
  treatment opens up new application domains and new possibilities for
  learning effectively from data.
  We validate the model on a referential 
  expression generation task, 
  showing that the best performance is
  achieved by incorporating features approximating well-established
  insights about natural language generation into RSA.
\end{abstract}
%


\section{Pragmatic language use}

In the Gricean view of language use \cite{Grice75}, people are
rational agents who are able to communicate efficiently and
effectively by reasoning in terms of shared communicative goals, the
costs of production, prior expectations, and others'
belief states. The Rational Speech Acts (RSA) model
\newcite{Frank:Goodman:2012} is a recent Bayesian reconstruction of
these core Gricean ideas.  RSA and its extensions have been shown to
capture many kinds of conversational implicature and to
closely model psycholinguistic data from children and adults
\cite{Degen:Franke:2012,Bergen:Levy:Goodman:2014,Kao-etal:2014,Potts-etal:2015,Stiller:Goodman:Frank:2011}.

Both Grice's theories and RSA have been criticized for predicting that people are
more rational than they actually are. These criticisms have been
especially forceful in the context of language production.  It seems
that speakers often fall short: their utterances are longer than they
need to be, underinformative, unintentionally ambiguous, obscure,
and so forth
\cite{baumann2014,engelhardt2006,Gatt-etal:2013b,levelt1993,McMahan:Stone:2015,pechmann1989}.
RSA can incorporate notions of bounded rationality
\cite{CamererHo:2004,Franke09DISS,Jaeger:2011}, but it still sharply
contrasts with views in the tradition of \cite{Dale:Reiter:1995},
in which speaker agents rely on heuristics and shortcuts to try to
accurately describe the world while managing the cognitive demands of
language production.

\newcommand{\ourModel}{learned RSA}
\newcommand{\OurModel}{Learned RSA}

In this paper, we offer a substantially different perspective on RSA
by showing how to define it as a trained statistical classifier,
which we call \tech{\ourModel}. At the heart of \ourModel\ is the
back-and-forth reasoning between speakers and listeners that
characterizes RSA.  However, whereas standard RSA requires a
hand-built lexicon, \ourModel\ infers a lexicon from data. And whereas
standard RSA makes predictions according to a fixed calculation,
\ourModel\ seeks to optimize the likelihood of whatever examples it is
trained on. Agents trained in this way exhibit the pragmatic behavior
characteristic of RSA, but their behavior is governed by their
training data and hence is only as rational as that experience
supports. To the extent that the speakers who produced the data are
pragmatic, \ourModel\ discovers that; to the extent that their
behavior is governed by other factors, \ourModel\ picks up on that
too.  We validate the model on the task of \tech{attribute selection
  for referring expression generation} with a widely-used corpus of
referential descriptions (the TUNA corpus;
\cite{vandeemter-vandersluis-gatt:2006:INLG,Gatt-etal:2007}), showing
that it improves on heuristic-driven models and pure RSA by
synthesizing the best aspects of both.



\section{RSA as a speaker model}\label{sec:rsa}

\newcommand{\Lewis}{r_{1}}
\newcommand{\Grice}{r_{2}}
\newcommand{\Shannon}{r_{3}}
\newcommand{\Eilenberg}{r_{4}}
\newcommand{\glasses}{\word{glasses}}
\newcommand{\tie}{\word{tie}}
\newcommand{\beard}{\word{beard}}

\begin{figure*}[t]
  \centering
  \setlength{\arraycolsep}{2pt}
  \setlength{\tabcolsep}{2pt}
  \subfigure[Simple reference game.]{
    \centering
    \begin{tabular}[b]{@{} *{3}{c} @{}}  
      \includegraphics[height=0.7in]{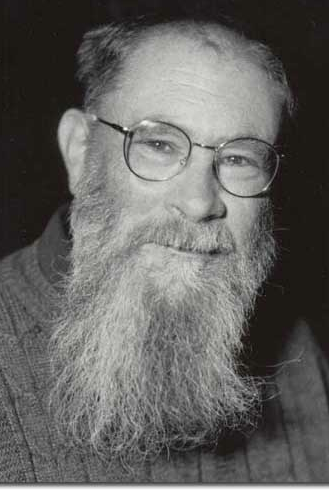} 
      &
        \includegraphics[height=0.7in]{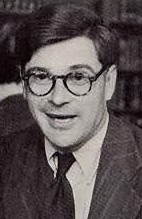} 
      & 
        \includegraphics[height=0.7in]{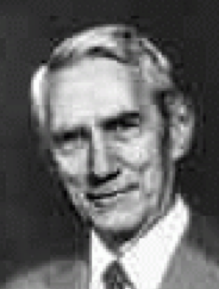} 
      \\[-0.5ex]
      $\Lewis$ & $\Grice$ & $\Shannon$
    \end{tabular}
    \label{fig:refgame}
  }
  \hfill
  \subfigure[$\speakerZero$]{\label{fig:s0}
    \centering
    $\begin{array}[b]{ rccc }
       \toprule
       & \uttheader{\beard} & \uttheader{\glasses} & \uttheader{\tie} \\
       \midrule  
       \Lewis   & .5 & .5 &  0 \\
       \Grice   &  0 & .5 & .5 \\
       \Shannon &  0 &  0 &  1  \\
       \bottomrule
    \end{array}$
  }
  \hfill
  \subfigure[$\listenerOne$]{
    \centering
    $\begin{array}[b]{ rccc }
       \toprule
       & \refheader{\Lewis} & \refheader{\Grice} & \refheader{\Shannon} \\
       \midrule                
       \beard   &   1 &  0  & 0\\
       \glasses &  .5 & .5  & 0 \\
       \tie     &   0 & .33 & .67 \\
       \bottomrule
    \end{array}$
  }
  \hfill
  \subfigure[$\speakerOne$]{
    \centering
    $\begin{array}[b]{ rccc }
       \toprule
       & \uttheader{\beard} & \uttheader{\glasses} & \uttheader{\tie} \\
       \midrule  
       \Lewis   & \graycell{.67} & .33 & 0 \\
       \Grice   &  0  & \graycell{.6} & .4 \\
       \Shannon &  0  &  0 &  \graycell{1} \\
       \bottomrule
     \end{array}$ 
  }
 \caption{Ambiguity avoidance in RSA.}
 \label{fig:rsa}
\end{figure*}

RSA is a descendent of the signaling systems of \cite{Lewis69} and
draws on ideas from iterated best response (IBR) models
\cite{Franke09DISS,Jaeger:2011}, iterated cautious response (ICR)
models \cite{Jaeger:2014}, and cognitive hierarchies
\cite{CamererHo:2004} (see also
\cite{Golland:Liang:Klein:2010,Rosenberg:Cohen:1964}).  RSA models
language use as a recursive process in which speakers and listeners
reason about each other to enrich the literal semantics of their
language.  This increases the efficiency and reliability of their
communication compared to what more purely literal agents can achieve.

For instance, suppose a speaker and listener are playing a reference
game in the context of the images in \figref{fig:refgame}. The speaker
$S$ has been privately assigned referent $r_{1}$ and must send a
message that conveys this to the listener. A literal speaker would
make a random choice between $\beard$ and $\glasses$. However, if $S$
places itself in the role of a listener $L$ receiving these messages,
then $S$ will see that $\glasses$ creates uncertainty about the
referent whereas $\beard$ does not, and so $S$ will favor $\beard$.
In short, the pragmatic speaker chooses $\beard$ because it's
unambiguous for the listener.

RSA formalizes this reasoning in probabilistic Bayesian terms. It assumes
a set of messages $\Messages$, a set of states $\States$, a prior
probability distribution $\StatePrior$ over states $\States$, and a
cost function $C$ mapping messages to real numbers. The semantics of
messages is defined by a lexicon~$\Lex$, where
$\Lex(\msg, \state) = 1$ if $\msg$ is true of $\state$ and $0$ otherwise.  The
agents are then defined as follows:
\begin{align}
  \label{s0}
  \speakerZero(\msg \given \state, \Lex) 
  &\propto 
  \exp\left({\lambda\left(\log\Lex(\msg, \state)- \Costs(\msg)\right)}\right) \\
  \label{l1}
  \listenerOne(\state \given \msg, \Lex) 
  &\propto 
  \speakerZero(\msg \given \state, \Lex) \StatePrior(\state) \\
%
  \label{s1}
  \speakerOne(\msg \given \state, \Lex) 
  &\propto 
  \exp\left({\lambda\left(\log\listenerOne(\state \given \msg, \Lex) - \Costs(\msg)\right)}\right)
\end{align}
The model that is the starting point for our contribution in this
paper is the pragmatic speaker $\speakerOne$. It reasons not about the
semantics directly but rather about a pragmatic listener
$\listenerOne$ reasoning about a literal speaker $\speakerZero$.  The
strength of this pragmatic reasoning is partly governed by the
temperature parameter $\lambda$, with higher values leading to more
aggressive pragmatic reasoning.

\Figref{fig:rsa} tracks the RSA computations for the 
reference game in \figref{fig:refgame}. Here, the message costs
$\Costs$ are all $0$, the prior over referents is flat, and
$\lambda=1$. The chances of success for the literal speaker
$\speakerZero$ are low, since it chooses true messages at random. In
contrast, the chances of success for $\speakerOne$ are high, since 
it derives the unambiguous system highlighted in gray.

The task we seek to model is a language generation task, so we
present RSA from a speaker-centric perspective. It has been explored
more fully from a listener perspective. In that
formulation, the model begins with a literal listener reasoning only
in terms of the lexicon $\Lex$ and state priors. Models of this
general form have been shown to capture a wide range of pragmatic
behaviors \cite{Bergen:Levy:Goodman:2014,Frank:Goodman:2014,Kao:Bergen:Goodman:2014,Kao-etal:2014,Potts-etal:2015}
and to increase success in
task-oriented dialogues \cite{Vogel-etal:2013,Vogel-etal:2014}.

RSA has been criticized on the grounds that it
predicts unrealistic speaker behavior \cite{Gatt-etal:2013b}.
For instance, in \figref{fig:rsa}, we confined our agents to a
simple message space. If permitted to use natural language, they
will often produce utterances expressing predicates that are redundant
from an RSA perspective---for example, by describing
$r_{1}$ as \word{the man with the long beard and sweater},
even though \word{man} has no power to discriminate, and \word{beard}
and \word{sweater} each uniquely identify the intended referent. This
tendency has several explanations, including a preference for
including certain kinds of descriptors, a desire to hedge against the
possibility that the listener is not pragmatic, and cognitive
pressures that make optimal descriptions impossible. One of our 
central objectives is to allow these factors to guide the core
RSA calculation.


\section{The TUNA corpus}\label{sec:tuna}

\begin{figure}[tp]
  \centering
  
\framebox{%
\scriptsize%
\setlength{\tabcolsep}{2pt}%
\begin{tabular}[c]{@{} *{3}{c} @{}}
\framebox{\parbox[c]{1.4cm}{\includegraphics[scale=0.08]{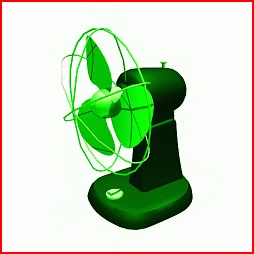}}
\begin{tabular}{@{} c @{}}
\refattr{colour:green}\\
\refattr{orientation:left}\\
\refattr{size:small}\\
\refattr{type:fan}\\
\refattr{x-dimension:1}\\
\refattr{y-dimension:1}
\end{tabular}} & \framebox{\parbox[c]{1.4cm}{\includegraphics[scale=0.08]{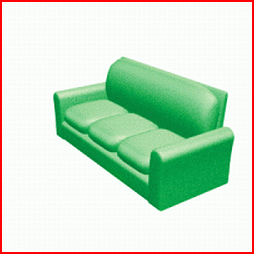}}
\begin{tabular}{@{} c @{}}
\refattr{colour:green}\\
\refattr{orientation:left}\\
\refattr{size:small}\\
\refattr{type:sofa}\\
\refattr{x-dimension:1}\\
\refattr{y-dimension:2}
\end{tabular}} & \framebox{\parbox[c]{1.4cm}{\includegraphics[scale=0.08]{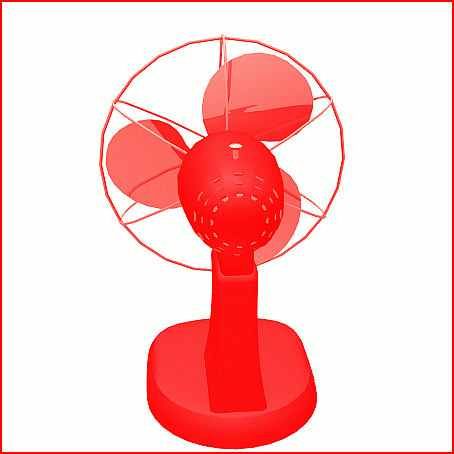}}
\begin{tabular}{@{} c @{}}
\refattr{colour:red}\\
\refattr{orientation:back}\\
\refattr{size:large}\\
\refattr{type:fan}\\
\refattr{x-dimension:1}\\
\refattr{y-dimension:3}
\end{tabular}}\\\\
\framebox{\parbox[c]{1.4cm}{\includegraphics[scale=0.08]{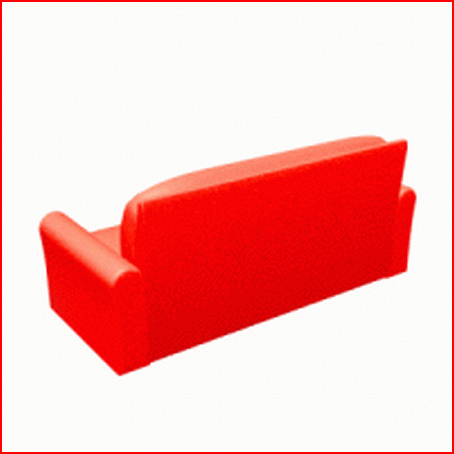}}
\begin{tabular}{@{} c @{}}
\refattr{colour:red}\\
\refattr{orientation:back}\\
\refattr{size:large}\\
\refattr{type:sofa}\\
\refattr{x-dimension:2}\\
\refattr{y-dimension:1}
\end{tabular}} & \framebox{\parbox[c]{1.4cm}{\includegraphics[scale=0.08]{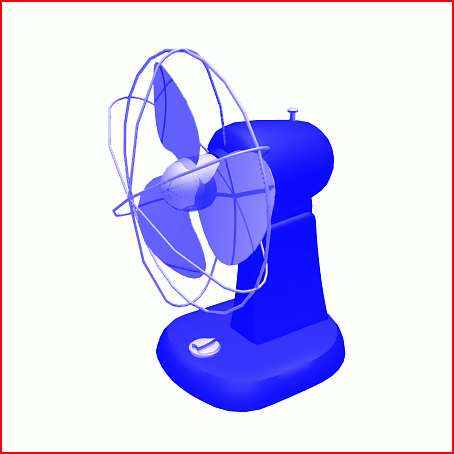}}
\begin{tabular}{@{} c @{}}
\refattr{colour:blue}\\
\refattr{orientation:left}\\
\refattr{size:large}\\
\refattr{type:fan}\\
\refattr{x-dimension:2}\\
\refattr{y-dimension:2}
\end{tabular}} & \\\\
\framebox{\parbox[c]{1.4cm}{\includegraphics[scale=0.08]{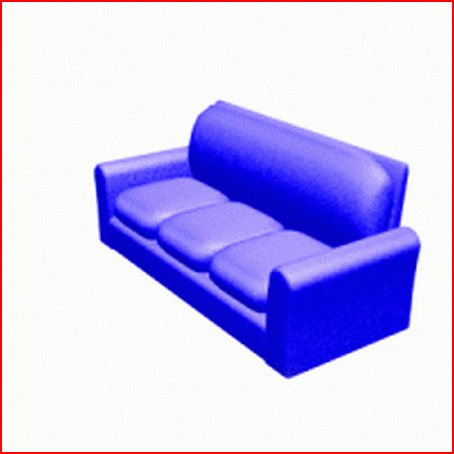}}
\begin{tabular}{@{} c @{}}
\refattr{colour:blue}\\
\refattr{orientation:left}\\
\refattr{size:large}\\
\refattr{type:sofa}\\
\refattr{x-dimension:3}\\
\refattr{y-dimension:1}
\end{tabular}} &  & \colorbox{lightgray}{\parbox[c]{1.4cm}{\includegraphics[scale=0.08]{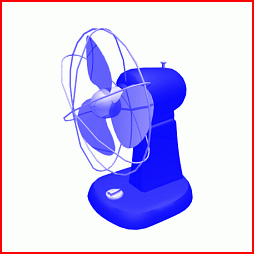}}
\begin{tabular}{@{} c @{}}
\refattr{colour:blue}\\
\refattr{orientation:left}\\
\refattr{size:small}\\
\refattr{type:fan}\\
\refattr{x-dimension:3}\\
\refattr{y-dimension:3}
\end{tabular}}
\end{tabular}
}

\vspace{10pt}

\footnotesize
\begin{tabular}{r@{: \ } l}Utterance &``blue fan small''\\
Utterance attributes &  \uttattr{colour:blue}; \uttattr{size:small}; \uttattr{type:fan}
\end{tabular}

  \caption{Example item from the TUNA corpus. Target is in gray.}
  \label{fig:TUNA}
\end{figure}

In \secref{sec:experiments}, we evaluate RSA and \ourModel\ in the
TUNA corpus
\cite{vandeemter-vandersluis-gatt:2006:INLG,Gatt-etal:2007}, a widely
used resource for developing and testing models of natural language
generation. We introduce the corpus now because doing so helps clarify
the learning task faced by our model, which we define in the next
section.

In the TUNA corpus, participants were assigned a target referent or
referents in the context of seven other distractors and asked to
describe the target(s). Trials were performed in two domains,
\cond{furniture} and \cond{people}, each with a \cond{singular}
condition (describe a single entity) and a \cond{plural} condition
(describe two). \Figref{fig:TUNA} provides a (slightly simplified)
example from the singular furniture section, with the target item
identified by shading. In this case, the participant wrote the message
``blue fan small''. All entities and messages are annotated with their
semantic attributes, as given in simplified form here.  (Participants
saw just the images; we include the attributes in \figref{fig:TUNA}
for reference.)

The task we address is \tech{attribute selection}: reproducing the
multiset of attributes in the message produced in each context. Thus,
for \figref{fig:TUNA}, we would aim to produce $\{$\uttattr{size:small},
\uttattr{colour:blue}, \uttattr{type:fan}$\}$. This is less demanding than full natural
language generation, since it factors out all morphosyntactic
phenomena. \Secref{sec:experiments} provides additional details on the
nature of this evaluation.


\section{\OurModel}\label{sec:ourmodel}

We now formulate RSA as a machine learning model
that can incorporate the quirks and limitations that characterize
natural descriptions while still presenting a unified model of
pragmatic reasoning. This approach builds on the two-layer
speaker-centric classifier of \cite{Golland:Liang:Klein:2010},
but differs from theirs in that we directly optimize the performance of the
pragmatic speaker in training, whereas \cite{Golland:Liang:Klein:2010}
apply a recursive reasoning model on top of a pre-trained classifier.
Like RSA, the model can be generalized to
allow for additional intermediate agents, and
it can easily be reformulated to begin with a literal listener.


\marginnote{Commented out the paragraph citing space limitations for the
denseness of our presentation. We can put it back in if they do end up
limiting our space, but it seems like a hollow excuse at the moment.}

\paragraph{Feature representations.}
To build an agent that learns effectively from data,
we must represent the items in our dataset in a
\marginnote{Changed instances of ``data set'' to ``dataset''
(both seem to be established spellings, so feel free to switch it to
two words as long as we're consistent).}
way that accurately captures their important distinguishing properties
and permits robust generalization to new items
\cite{Domingos:2012,Liang:Potts:2015}.  We define our feature
representation function $\phi$ very generally as a map from
state--utterance--context triples $\tuple{\state, \msg, \context}$ to
vectors of real numbers.  This gives us the freedom to design the
feature function to encode as much relevant information as necessary.

As noted above, in \ourModel, we do not presuppose a semantic lexicon,
but rather induce one from the data as part of learning.  The feature
representation function determines
a large, messy hypothesis space of potential lexica that is refined
during optimization. For instance, as a starting point, we might
define the feature space in terms of the cross-product of all possible
entity attributes and all possible utterance meaning attributes. For
$m$ entity attributes and $n$ utterance attributes, this defines each
$\phi(\state, \msg, \context)$ as an $mn$-dimensional vector. Each
dimension of this vector records the number of times that its
corresponding pair of attributes co-occurs in $\state$ and $\msg$.
Thus, the representation of the target entity in
\figref{fig:TUNA} would include a $1$ in the dimension for clearly
good pairs like \featpair{colour:blue}{colour:blue} as well as for intuitively
incorrect pairs like \featpair{colour:blue}{size:small}.

Because $\phi$ is defined very generally, we can also include
information that is not clearly lexical. For instance, in our
experiments, we add dimensions that count the color
attributes in the utterance in various ways, ignoring the specific
color values. We can also define features that intuitively
involve negation, for instance, those that capture entity attributes
that go unmentioned. This freedom is crucial to bringing
generation-specific insights into the RSA reasoning.

\paragraph{Literal speaker.}

\newcommand{\Szero}{S_{0}}
\newcommand{\PSzero}{\Szero}
\newcommand{\Lone}{L_{1}}
\newcommand{\Sone}{S_{1}}

\OurModel\ is built on top of a \tech{log-linear
  model}, standard in the machine learning literature and widely
applied to classification tasks
\cite{Hastie:Tibshirani:Friedman:2009,McCullagh:Nelder:1989}.
\begin{equation}
  \label{S0}
  \PSzero(\msg \given \state, \context; \theta) \propto
    \exp(\theta^{T} \phi(\state, \msg, \context))
\end{equation}
This model serves as our literal speaker, analogous to $\speakerZero$
in \eqref{s0}.  The lexicon of this model is embedded in the
\tech{parameters} (or \tech{weights}) $\theta$.  Intuitively, 
$\theta$ is the direction in
feature representation space that the literal speaker believes is most
positively correlated with the probability that the message will be
correct.  We train the model by searching for a $\theta$ to maximize
the conditional likelihood the model assigns to the messages in the training examples. Assuming the training is
effective, this increases the weight for correct pairings between
utterance attributes and entity attributes and decreases the weight
for incorrect pairings. 

To find the optimal $\theta$, we seek to maximize the conditional likelihood of the training examples
using first-order optimization methods (described in more detail in
\pararef{Learning}, below). This requires the gradient of the likelihood
with respect to $\theta$. To simplify the gradient derivation and
improve numerical stability, we maximize the log of the conditional
likelihood:
\begin{equation}
  \label{grad:Szero}
  J_{\Szero}(\state, \msg, \context, \theta) = \log \PSzero(\msg \given \state, \context; \theta)
\end{equation}
The gradient of this log-likelihood is
\begin{eqnarray}
  \frac{\partial J_{\Szero}}{\partial \theta} 
  &=& \phi(\state, \msg, \context) - \frac{1}{\sum_{\msg'} \exp(\theta^{T} \phi(\state, \msg', \context))} \sum_{\msg'} \exp(\theta^{T} \phi(\state, \msg', \context)) \phi(\state, \msg', \context) \notag\\
  &=& \phi(\state, \msg, \context) - \sum_{\msg'} \PSzero(\msg' \given \state, \context; \theta) \phi(\state, \msg', \context) \notag\\
  &=& \phi(\state, \msg, \context) - \expect_{\msg' \sim \PSzero(\cdot \given \state, \context; \theta)}  \left[\phi(\state, \msg', \context)\right] \label{eqn:literal}
\end{eqnarray}
where the first two equations can be derived by expanding the proportionality constant in the definition of $\Szero$.

\paragraph{Pragmatic speaker.}

We now define a pragmatic listener $\Lone$ and a pragmatic speaker
$\Sone$. We will show experimentally (\secref{sec:experiments}) that
the learned pragmatic speaker $\Sone$ agrees better with human speakers
on a referential expression generation task than either the literal speaker
$\Szero$ or the pure RSA speaker $\speakerOne$.

The parameters for
$\Lone$ and $\Sone$ are still the parameters of the
literal speaker $\Szero$; we wish to update them to maximize the
performance of $\Sone$, the agent that acts according to
$\Sone(\msg \given \state, \context; \theta)$, where
\begin{align}
  \Sone(\msg \given \state, \context; \theta) &\propto
    \Lone(\state \given \msg, \context; \theta) \\
  \Lone(\state \given \msg, \context; \theta) &\propto
    \Szero(\msg \given \state, \context; \theta)
\end{align}
This corresponds to the simplest case of RSA in which $\lambda = 1$
and message costs and state priors are uniform:
$\speakerOne(\msg \given\state,\Lex) \propto \listenerOne(\state \given
\msg,\Lex) \propto \speakerZero(\msg \given\state,\Lex)$.

\marginnote{Added this paragraph to connect with the abstract. The
citation is a bit arbitrary (an early and famous neural net paper
that uses the terminology ``activation function'').}

In optimizing the performance of the pragmatic speaker $\Sone$ by adjusting
the parameters to the simpler classifier $\Szero$, the RSA back-and-forth
reasoning can be thought of as a non-linear function through which
errors are propagated in training, similar to the activation functions
in neural network models \cite{Rumelhart-etal:1985}. However, unlike neural
network activation functions, the RSA reasoning applies a different
non-linear transformation depending on the pragmatic context
(sets of available referents and utterances).

For convenience, we define symbols for the log-likelihood of each of
these probability distributions:
\begin{align}
  J_{\Sone}(\state, \msg, \context, \theta) &= \log \Sone(\msg \given \state, \context; \theta) \\
  J_{\Lone}(\state, \msg, \context, \theta) &= \log \Lone(\state \given \msg, \context; \theta)
\end{align}

The log-likelihood of each agent has the same form as the
log-likelihood of the literal speaker, but with the value of the
distribution from the lower-level agent substituted for the score
$\theta^T \phi$. By a derivation similar to the one in
\eqref{eqn:literal} above, the gradient of these log-likelihoods can
thus be shown to have the same form as the gradient of the literal
speaker, but with the gradient of the next lower agent substituted for
the feature values:
\begin{align}
  \frac{\partial J_{\Sone}}{\partial \theta} &= \frac{\partial J_{\Lone}}{\partial \theta}(\state, \msg, \context, \theta) - \expect_{\msg' \sim \Sone(\cdot \given \state, \context; \theta)} \left[\frac{\partial J_{\Lone}}{\partial \theta}(\state, \msg', \context, \theta)\right] \label{grad:Sone} \\
  \frac{\partial J_{\Lone}}{\partial \theta} &= \frac{\partial J_{\Szero}}{\partial \theta}(\state, \msg, \context, \theta) - \expect_{\state' \sim \Lone(\cdot \given \msg, \context; \theta)} \left[\frac{\partial J_{\Szero}}{\partial \theta}(\state', \msg, \context, \theta)\right] \label{grad:Lone}
\end{align}
The value $J_{\Szero}$ in \eqref{grad:Lone} is as defined in
\eqref{grad:Szero}.

\paragraph{Training.}

As mentioned above, our primary objective in training is to maximize
the (log) conditional likelihood of the messages in the training examples
given their respective states and contexts. We add to this an $\ell_{2}$
regularization term, which expresses a Gaussian prior distribution over
the parameters $\theta$. Imposing this prior helps prevent overfitting to the
training data and thereby damaging our ability to generalize well to new
examples \cite{Chen:Rosenfeld:1999}.
With this modification, we instead maximize the log of the
posterior probability of the parameters and the training examples
jointly. For a dataset of $M$ training examples
$\tuple{\state_i, \msg_i, \context_i}$, this log posterior is:
\begin{equation}
J(\theta) = -\frac{M}{2}\ell||\theta||^2 + \sum_{i=1}^M \log S_1(\msg_{i} \given \state_{i}, \context_{i}; \theta)
\end{equation}

The stochastic gradient descent (SGD) family of first-order optimization
techniques \cite{Bottou:2010} can be used to approximately maximize $J(\theta)$
by obtaining noisy estimates of its gradient
and ``hill-climbing'' in the
direction of the estimates. (Strictly speaking, we are employing
stochastic gradient \emph{ascent} to maximize the objective rather than
minimize it; however, SGD is the much more commonly seen term for the technique.)

The exact gradient of this objective function is
\begin{equation}
\frac{\partial J}{\partial \theta} = -M\ell \theta + \sum_{i=1}^M \frac{\partial J_{\Sone}}{\partial \theta}(\state_{i}, \msg_{i}, \context_{i}, \theta)
\end{equation}
using the per-example gradient
$\frac{d J_{\Sone}}{d \theta}$ given in \eqref{grad:Sone}.
SGD uses the per-example gradients (and a simple scaling of the $\ell_2$
regularization penalty) as its noisy estimates, thus relying on each example
to guide the model in roughly the correct direction towards the optimal
parameter setting. Formally, for each example $(\state, \msg, \context)$,
the parameters are updated according to the formula
\begin{eqnarray}
\theta &:=& \theta + \alpha \left(-\ell\theta + \frac{\partial J_{\Sone}}{\partial \theta}(\state, \msg, \context, \theta)\right) 
\end{eqnarray}

The learning rate $\alpha$ determines how ``aggressively'' the parameters
are adjusted in the direction of the gradient. Small values of $\alpha$ lead
to slower learning, but a value of $\alpha$ that is too large can result in
the parameters overshooting the optimal value and diverging.
To find a good learning rate, we use AdaGrad \cite{Duchi:Hazan:Singer:2011},
which sets the learning rate
adaptively for each example based on an initial step size $\eta$ and gradient history.
The effect of AdaGrad is to reduce the learning rate over time such that
the parameters can settle down to a local optimum despite the noisy gradient
estimates, while continuing to allow high-magnitude updates along certain
dimensions if those dimensions have exhibited less noisy behavior in previous
updates.


\section{Example}

\begin{figure}[htp]
  \centering
  \setlength{\arraycolsep}{2pt}
  \setlength{\tabcolsep}{2pt}
  \subfigure[Learned $\Sone$ model training. Gradient values given are $6 \frac{\partial J_{\Sone}}{\partial \theta}$, evaluated at $\theta = \vec{0}$.]{
    \centering
    \begin{tabular}[b]{@{} *{1}{r} *{2}{c} @{\hskip 0.5em} *{2}{c} @{\hskip 0.5em} *{1}{c} @{}}
    	  & \multicolumn{3}{c}{Training examples} & & Test example
    	  \\
    	  \\
    	  Context
    	  &
        \begin{tabular}{cc}
          \fbox{\includegraphics[height=0.7in]{fig/Grice}}
        & 
          \includegraphics[height=0.7in]{fig/Shannon}
        \\
          $\Grice$ & $\Shannon$
        \end{tabular}
      & &
      	  \begin{tabular}{cc}
          \includegraphics[height=0.7in]{fig/Shannon}
        &
          \fbox{\includegraphics[height=0.7in]{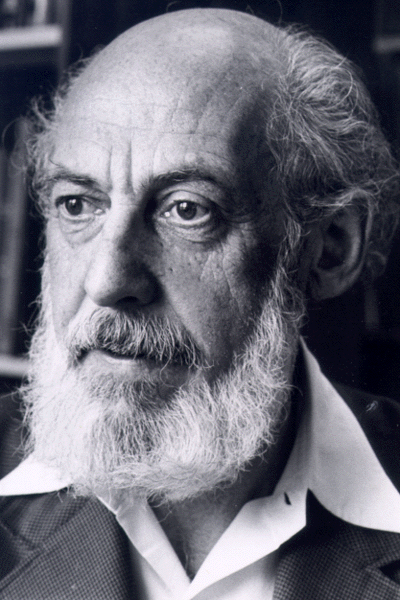}}
        \\
          $\Shannon$ & $\Eilenberg$
    	    \end{tabular}
      & &
    	    \begin{tabular}{cc}
          \fbox{\includegraphics[height=0.7in]{fig/Lewis}}
        &
          \includegraphics[height=0.7in]{fig/Eilenberg}
        \\
          $\Lewis$ & $\Eilenberg$
    	    \end{tabular}
      \\
      \\
      Utterance
      &
      	  \word{\uttattr{person} with \uttattr{glasses}}
      & &
        \word{\uttattr{person} with \uttattr{beard}}
      & &
        \word{\uttattr{person} with \uttattr{glasses}}
      \\
      \\
      \begin{tabular}{r}Features \\ for true \\ utterance\end{tabular}
      &
      \begin{tabular}{c}
	    \featpair{person}{person} \\
        \featpair{glasses}{person} \\
        \featpair{person}{glasses} \\
        \featpair{glasses}{glasses} \\
        \featpair{person}{$\neg$beard} \\
        \featpair{glasses}{$\neg$beard} \\
      \end{tabular}
      & &
      \begin{tabular}{c}
        \featpair{person}{person} \\
        \featpair{beard}{person} \\
        \featpair{person}{$\neg$glasses} \\
        \featpair{beard}{$\neg$glasses} \\
        \featpair{person}{beard} \\
        \featpair{beard}{beard} \\
      \end{tabular}
      & &
      \begin{tabular}{c}
        \featpair{person}{person} \\
        \featpair{glasses}{person} \\
        \featpair{person}{glasses} \\
        \featpair{glasses}{glasses} \\
        \featpair{person}{beard} \\
        \featpair{glasses}{beard} \\
      \end{tabular}
      \\
      \\
      \begin{tabular}{r}
        Gradient
      \end{tabular}
      &
      \begin{tabular}{ rrrr }
         \toprule
         &
           \uttheader{\uttattr{person}} &
           \uttheader{\uttattr{glasses}} &
           \uttheader{\uttattr{beard}} \\
         \midrule  
         \refattr{person}        & \sigweight{1} & 1 &  -1 \\
         \refattr{glasses}       & \sigweight{2} & \sigweight{2} & \sigweight{-2} \\
         \refattr{beard}         &  0 &  0 &  0 \\
         \refattr{$\neg$glasses} & -1 & \sigweight{-1} & \sigweight{1} \\
         \refattr{$\neg$beard}   &  1 & \sigweight{1} & \sigweight{-1} \\
         \bottomrule
      \end{tabular}
      & &
      \begin{tabular}{ rrrr }
         \toprule
         &
           \uttheader{\uttattr{person}} &
           \uttheader{\uttattr{glasses}} &
           \uttheader{\uttattr{beard}} \\
         \midrule  
         \refattr{person}        & \sigweight{1} & -1 &  1 \\
         \refattr{glasses}       &  0 &  0 &  0 \\
         \refattr{beard}         & \sigweight{2} & \sigweight{-2} & \sigweight{2} \\
         \refattr{$\neg$glasses} &  1 & \sigweight{-1} & \sigweight{1} \\
         \refattr{$\neg$beard}   & -1 & \sigweight{1} & \sigweight{-1} \\
         \bottomrule
      \end{tabular}
      & &
      (unused)
      \\
      \\
    \end{tabular}
    \label{fig:traindata}
  } \\
  \subfigure[Pure RSA ($\speakerOne$), linear classifier ($\Szero$), and learned RSA ($\Sone$) utterance distributions. RSA alone minimizes ambiguity but can't learn overgeneration from the examples. The linear classifier learns to produce \uttattr{person} but fails to minimize ambiguity. The weights in \ourModel\ retain the tendency to produce \uttattr{person} in all cases, while the recursive reasoning yields a preference for the unambiguous descriptor \uttattr{glasses}.]{
    \centering
    \qquad\qquad\qquad
    $\begin{array}[b]{ ccc @{\hskip 1em} ccc @{\hskip 1em} ccc @{\hskip 1em} l }
       \toprule
       \multicolumn{2}{c}{\speakerOne} & &
       \multicolumn{2}{c}{\Szero} & &
       \multicolumn{2}{c}{\Sone}
       \\
         \refheader{\Lewis}
       &
         \refheader{\Eilenberg}
       & &
         \refheader{\Lewis}
       &
         \refheader{\Eilenberg}
       & &
         \refheader{\Lewis}
       &
         \refheader{\Eilenberg}
       \\
       \midrule  
       .08 & \graycell{.25} & &
       .03 & .00 & &
       .10 & .11 & &
       \emptystring \\
       .08 & \graycell{.25} & &
       \graycell{.22} & .10 & &
       .16 & .13 & &
       \word{\uttattr{person}} \\
       \graycell{.17} & 0 & &
       .03 & .00 & &
       .11 & .07 & &
       \word{\uttattr{glasses}} \\
       .08 & \graycell{.25} & &
       .03 & .04 & &
       .08 & .17 & &
       \word{\uttattr{beard}} \\
       \graycell{.17} & 0 & &
       \graycell{.22} & .01 & &
       \graycell{.18} & .08 & &
       \word{\uttattr{person}, \uttattr{glasses}} \\
       .08 & \graycell{.25} & &
       \graycell{.22} & \graycell{.74} & & 
       .12 & \graycell{.19} & &
       \word{\uttattr{person}, \uttattr{beard}} \\
       \graycell{.17} & 0 & &
       .03 & .00 & &
       .10 & .11 & &
       \word{\uttattr{glasses}, \uttattr{beard}} \\
       \graycell{.17} & 0 & &
       \graycell{.22} & .10 & &
       .16 & .11 & &
       \word{\uttattr{person}, \uttattr{glasses}, \uttattr{beard}} \\
       \bottomrule
     \end{array}$
     \qquad\qquad\qquad
     \label{fig:distributions}
  }
 \caption{Specificity implicature and overgeneration in learned RSA.}
 \label{fig:learning}
\end{figure}

In \figref{fig:learning}, we illustrate crucial aspects of how our
model is optimized, fleshing out the concepts from the previous
section. The example also shows the ability of the trained $\Sone$
model to make a specificity implicature without having observed one in
its data, while preserving the ability to produce uninformative
attributes if encouraged to do so by experience.


As in our main experiments, we frame the learning task in terms of
attribute selection with TUNA-like data.  In this toy experiment, the
agent is trained on two example contexts, consisting of a target
referent, a distractor referent, and a human-produced utterance.  It
is evaluated on a third test example. This small dataset is given in
the top two rows of \figref{fig:learning}. The utterance on the
test example is shown for comparison; it is not provided to
the agent.

Our feature representations of the data are in the third row.
Attributes of the referents are in \refattr{small
  caps}; semantic attributes of the utterances are in
\uttattr{square brackets}.  These representations employ the 
cross-product features described in \secref{sec:ourmodel};
in TUNA data, properties that the target entities do not possess
(e.g., \refattr{$\neg$glasses}) are also included among their 
``attributes.''

Below the feature representations, we summarize the gradient of the
log likelihood ($\frac{\partial J_{\Sone}}{\partial \theta}$) for each example,
as an $m \times n$ table
representing the weight update for each of the $mn$ cross-product features.
(We leave out the $\ell_2$ regularization and AdaGrad learning rate
for simplicity.) Tracing the formula for this gradient \eqref{grad:Sone} back through the RSA layers to the literal listener
\eqref{grad:Szero}, one can see that the gradient consists of the feature
representation of the triple $\tuple{\state, \msg, \context}$ containing the
correct (human-produced) message, minus adjustments that penalize the
other messages according to how much the model was ``fooled'' into expecting them.

The RSA reasoning yields gradients that express both lexical and contextual
knowledge. From the first training example, the model learns the lexical
information that \uttattr{person} and \uttattr{glasses} should be used to
describe the target. However, this knowledge receives
higher weight in the association with \refattr{glasses},
because that attribute is disambiguating in this context.
As one would hope, the overall result is that intuitively good pairings
generally have higher weights, though the training set is too small to
fully distinguish good features from bad ones. For example, after seeing
both training examples and failing to observe both a beard and glasses on
the same individual, the model incorrectly infers that \uttattr{beard}
can be used to indicate a lack of glasses and vice versa. Additional
training examples could easily correct this.

\Figref{fig:distributions} shows the distribution over utterances given
target referent as predicted by
the learned pragmatic speaker $\Sone$ after one pass through the data
with a fixed learning rate $\alpha = 1$ and no regularization ($\ell = 0$).
We compare this distribution with the distribution predicted by the
learned literal speaker $\Szero$ and the pure RSA speaker $\speakerOne$.
We wish to determine whether each model can \begin{enumerate*}[label=(\roman*)]
\item minimize ambiguity; and
\item learn a prior preference for producing certain descriptors 
even if they are redundant.
\end{enumerate*}

The distributions in \figref{fig:distributions} show that
the linear classifier correctly learns that human-produced utterances in the
training data tend to mention the attribute \uttattr{person} even though
it is uninformative. However, for the referent that was not seen in the
training data, the model cannot decide among mentioning \uttattr{beard},
\uttattr{glasses}, both, or neither, even though the messages that don't
mention \uttattr{glasses} are ambiguous in context. The pure RSA model,
meanwhile, chooses messages that are unambiguous, but because it
has no mechanism for learning from the examples, it does not prefer to
produce \uttattr{person} without a manually-specified prior.

Our pragmatic speaker $\Sone$ gives us the best of both models: the parameters
$\theta$ in \ourModel\ show the tendency exhibited in the training
data to produce \uttattr{person} in all cases, while the RSA
recursive reasoning mechanism guides the model to produce unambiguous
messages by including the attribute \uttattr{glasses}.

\section{Experiments}
\label{sec:experiments}

\paragraph{Data.}
We report experiments on the TUNA corpus (\secref{sec:tuna} above).
We focus on the \cond{singular} portion of the corpus, which was used
in the 2008 and 2009 Referring Expression Generation Challenges. 
We do not have access to the train/dev/test
splits from those challenges, so we report five-fold
cross-validation numbers. The singular portion consists of 420
\cond{furniture} trials involving 176 distinct referents and 360 \cond{people}
trials involving 228 distinct referents.


\paragraph{Evaluation metrics.}

\newcommand{\Dice}{\emph{Dice}}
\newcommand{\Count}[2]{\mathbb{Z}_{#2}{(#1)}}
\newcommand{\attr}{a}

The primary evaluation metric used in the attribute selection task
with TUNA data is \tech{multiset Dice} calculated
on the attributes of the generated messages:
\begin{equation}
  \label{dice}
  \frac{
    2\sum_{x \in D} \min\left[\Count{x}{\attr(\msg_{i})}, \Count{x}{\attr(\msg_{j})}\right]
    }{
      |\attr(\msg_{i})| +  |\attr(\msg_{j})|
    }
\end{equation}
Here, $\attr(\msg)$ is the multiset of attributes of message $\msg$,
$D$ is the non-multiset union of $\attr(\msg_{i})$ and
$\attr(\msg_{j})$, $\Count{x}{X}$ is the number of occurrences of $x$
in the multiset $X$, and $|\attr(\msg_{i})|$ is the cardinality of
multiset $\attr(\msg)$.  \tech{Accuracy} is the fraction of examples
for which the subset of attributes is predicted perfectly (equivalent to achieving
multiset Dice $1$).

\paragraph{Experimental set-up.}

We evaluate all our agents in the same pragmatic contexts: for each
trial in the \cond{singular} corpus, we define the messages $\Messages$ to be
the powerset of the attributes used in the referential description and
the states $\States$ to be the set of entities in the trial, including
the target. The message predicted by a speaker agent is the one
with the highest probability given the target entity; if more than one
message has the highest probability, we allow the agent to choose
randomly from the highest probability ones.

In learning,
we use initial step size $\eta = 0.01$ and regularization constant
$\ell = 0.01$. RSA agents are not trained, but we cross-validate to optimize
$\lambda$ and the function defining message costs,
choosing from
\begin{enumerate*}[label=(\roman*)]
\item $\Costs(\msg) = 0$;
\item $\Costs(\msg) = |\attr(\msg)|$; and
\item $\Costs(\msg) = -|\attr(\msg)|$.
\end{enumerate*}

\paragraph{Features.}
\marginnote{Swapped the ``gen.\ feats.\ only'' rows so we can save
the best for last :)}

We use \tech{indicator features} as our feature
representation; that is, the dimensions of the
feature representation take the values 0 and 1, with 1 representing the
truth of some predicate $P(\state, \msg, \context)$ and 0 representing its
negation. Thus, each vector of real numbers
that is the value of $\phi(\state, \msg, \context)$ can be represented
compactly as a set of predicates.

The baseline feature set consists of indicator features over all
conjunctions of an attribute of the referent and an attribute in 
the candidate message (e.g., $P(\state, \msg, \context) = \refattr{red}(\state) \wedge {}\uttattr{blue}{} \in \msg$).  We compare this to a version of the model
with additional \tech{generation features} that
seek to capture the preferences identified in prior work on
generation. These consist of indicators over the following features
of the message:
\begin{enumerate}[label=(\roman*)]
\item attribute type (e.g., $P(\state, \msg, \context) =$ ``$\msg$ contains a color'');
\item pair-wise attribute type co-occurrences, where one can be negated (e.g., ``$\msg$ contains a color and a size'', ``$\msg$ contains an object type but not a color''); and
\item message size in number of attributes (e.g., ``$\msg$ consists of 3 attributes'').
\end{enumerate}
For comparison, we also separately train literal speakers $\Szero$
as in \eqref{S0} (the log-linear model)
with each of these feature sets using the same optimization
procedure.

\begin{table*}[t]
  \centering
  \caption{Experimental results: mean accuracy and multiset Dice (five-fold cross-validation). \best{Bold}: best result; \close{bold italic}: not significantly different from best ($p >\;$0.05, Wilcoxon signed-rank test).}
  \begin{tabular}[t]{@{} l r@{\% \ }  r@{ \ }  r@{\% \ } r@{ \ }  r@{\% \ } r@{ \ } @{}}
    \toprule
          & \multicolumn{2}{c}{Furniture} & \multicolumn{2}{c}{People} & \multicolumn{2}{c}{All} \\
    Model & \multicolumn{1}{c}{Acc.} & \multicolumn{1}{c}{Dice} & \multicolumn{1}{c}{Acc.} & \multicolumn{1}{c}{Dice} & \multicolumn{1}{c}{Acc.} & \multicolumn{1}{c}{Dice} \\
    \midrule    
    RSA $\speakerZero$ (random true message)   & 1.0 & .475 & 0.6 & .125 & 1.7 & .314 \\
    RSA $\speakerOne$    & 1.9 & .522 &  2.5 & .254 & 2.2 & .386 
    \\[1ex]
    Learned $\Szero$, basic feats.       & 16.0 & .779 & 9.4 & .697 & 12.9 & .741 \\ 
    Learned $\Szero$, gen.\ feats.\ only & 5.0 & .788 & 7.8 & .681 & 6.3 & .738 \\ 
    Learned $\Szero$, basic $+$ gen.\ feats. & \best{28.1} & \best{.812} & 17.8 & .730 & \close{23.3} & \close{.774} 
    \\[1ex]
    Learned $\Sone$, basic feats.       & 23.1 & .789 & 11.9 & .740 & 17.9 & .766 \\ 
    Learned $\Sone$, gen.\ feats.\ only & 17.4 & .740 & 1.9 & .712 & 10.3 & .727 \\ 
    Learned $\Sone$, basic $+$ gen.\ feats. & \close{27.6} & .788 & \best{22.5} & \best{.764} & \best{25.3} & \best{.777} \\ 
    \bottomrule
  \end{tabular}
  \label{tab:results}
\end{table*}

\paragraph{Results.}

The results (\tabref{tab:results}) show that training a speaker agent
with learned RSA generally improves generation over the ordinary
classifier and RSA models. On the more complex
\cond{people} dataset, the pragmatic $\Sone$ model significantly outperforms all other
models. The value of the model's flexibility in allowing a variety of
feature designs can be seen in the comparison of the different feature
sets: we observe consistent gains from adding generation features to the
basic cross-product feature set. \marginnote{Added a sentence about the
significance of the ``gen.\ feats.\ only'' rows.}
Moreover, the two types of features complement
each other: neither the cross-product features nor the generation features in
isolation achieve the same performance as the combination of the two.

Of the models in \tabref{tab:results}, all but the last exhibit
systematic errors. Pure RSA performs poorly for reasons predicted by
\newcite{Gatt-etal:2013b}---for example, it under-produces color terms and head nouns like \word{desk}, \word{chair}, 
and \word{person}. This problem is also observed
in the trained $S_1$ model, but
is corrected by the generation features. On the \cond{people} dataset, the
$S_0$ models under-produce \word{beard} and \word{hair}, which
are highly informative in certain contexts. This type of communicative
failure is eliminated in the $S_1$ speakers.

The performance of the \ourModel\ model on the \cond{people} trials
also compares favorably to the best dev set performance numbers from
the 2008 Challenge \cite{Gatt-etal:2008:TUNA}, namely, .762 multiset
Dice, although this comparison must be informal since the test sets
are different. (In particular, the Accuracy values given in
\cite{Gatt-etal:2008:TUNA} are unfortunately not comparable with the
values we present, as they reflect ``perfect match with \emph{at least
  one} of the two reference outputs'' [emphasis in original].)
Together, these results show the value of being able to train a single
model that synthesizes RSA with prior work on generation.


\section{Conclusion}

Our initial experiments demonstrate the utility of RSA as a trained
classifier in generating referential expressions. The primary
advantages of this version of RSA stem from the flexible ways in which
it can learn from available data. This not only removes the need to
specify a complex semantic lexicon by hand, but it also provides the
analytic freedom to create models that are sensitive to factors
guiding natural language production that are not naturally expressed
in standard RSA.

This basic presentation suggests a range of potential next steps.  For
instance, it would be natural to apply the model to pragmatic
interpretation (the listener's perspective); this requires no
substantive formal changes to the model as defined in
\secref{sec:ourmodel}, and it opens up new avenues in terms of
evaluating pragmatic models in standard classification tasks like
sentiment analysis, topic prediction, and natural language reasoning.
In addition, for all versions of the model, one could consider
including additional hidden speaker and listener layers, incorporating
message costs and priors into learning, to capture a wider range of
pragmatic phenomena.


\bibliographystyle{plain}
\bibliography{monroe-potts-emnlp2015-bib}
\end{document}